%% file: root.tex
\documentclass[letterpaper, 10 pt, conference]{ieeeconf}  
\IEEEoverridecommandlockouts                              
\usepackage[noadjust]{cite}
\usepackage{graphicx}
\usepackage{amsmath}

\usepackage{multirow}
\usepackage[normalem]{ulem}
\usepackage{hyperref}
\hypersetup{
    colorlinks=true,
    linkcolor=blue,
    filecolor=magenta,      
    urlcolor=blue,
}

\useunder{\uline}{\ul}{}
\title{\LARGE \bf Supervision via Competition: Robot Adversaries for Learning Tasks}
\author{Lerrel Pinto$^{1}$, James Davidson$^{2}$ and Abhinav Gupta$^{1,3}$\\
$^{1}$ Carnegie Mellon University, $^{2}$ Google Brain, $^{3}$Google Research
}
\begin{document}
\maketitle
\thispagestyle{empty}
\pagestyle{empty}
\input{abstract.tex}
\input{introduction.tex}

\input{relatedWork.tex}
\input{approach.tex}

\input{results.tex}

\input{conclusion.tex}

\section*{ACKNOWLEDGEMENTS}
This work was supported by ONR MURI N000141612007, NSF IIS-1320083 and gift from Google.

\bibliographystyle{IEEEtran}
\bibliography{IEEEabrv,references}
\end{document}

%% file: abstract.tex
\begin{abstract}
There has been a recent paradigm shift in robotics to data-driven learning for planning and control. Due to large number of experiences required for training, most of these approaches use a self-supervised paradigm: using sensors to measure success/failure. However, in most cases, these sensors provide weak supervision at best. In this work, we propose an adversarial learning framework that pits an adversary against the robot learning the task. In an effort to defeat the adversary, the original robot learns to perform the task with more robustness leading to overall improved performance. We show that this adversarial framework forces the the robot to learn a better grasping model in order to overcome the adversary. By grasping 82\% of presented novel objects compared to 68\% without an adversary, we demonstrate the utility of creating adversaries. We also demonstrate via experiments that having robots in adversarial setting might be a better learning strategy as compared to having collaborative multiple robots.
\\
\noindent For supplementary video see: \url{youtu.be/QfK3Bqhc6Sk}


\end{abstract}

%% file: introduction.tex
\vspace{0.2in}
\section{INTRODUCTION}
There has been a recent push in robotics to move from analytical reasoning to more data-driven and self-supervised learning of planning and control. Specifically, the paradigm of end-to-end learning has gained a lot of prominence. In an end-to-end learning framework, the input is the perceived image and the output is the action primitives or torques themselves. Amazingly, it has been shown that given a specific task, if enough amount of data is collected, these frameworks can outperform manually designed mathematical models~\cite{lenz2015deep, pinto2016supersizing, levine2016end, levine2016learning}.

\begin{figure}[t!]
\begin{center}
\includegraphics[width=3.3in]{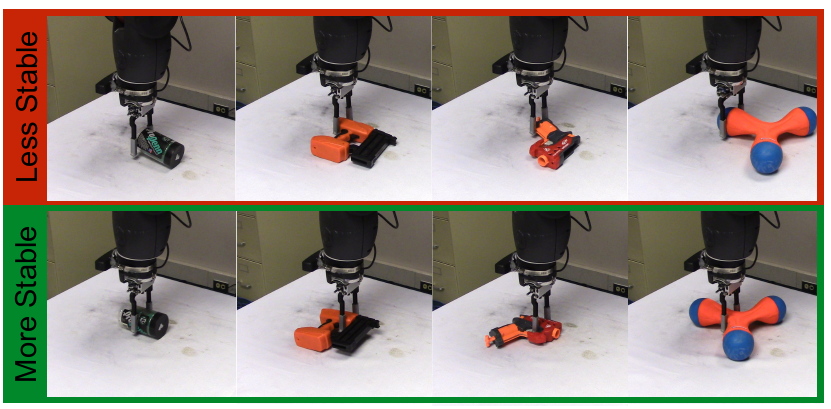}
\end{center}
\caption{ (first row) Examples of successful yet unstable grasps. (second row) Examples of stable and successful grasps.}
\label{fig:unstable}
\end{figure}

However, the critics of self-supervised learning often argue that the amount of data required to learn these models effectively is huge and presents a big bottleneck. For example, in our recent paper \cite{pinto2016supersizing}, we collect more than 50K examples to learn a grasping model. Furthermore, since most end-to-end learning approaches deal with large amounts of data, they use self-supervision from other sensors. This more often leads to weaker notion of success. For example, \cite{pinto2016supersizing} uses a force sensor in the gripper to see if the robot was successful in grasping the object. This often leads to an unstable grasp (see figure~\ref{fig:unstable} for examples) being classified as a good grasp in the self-supervised framework. 

To address the data scalability concerns, researchers at Google developed an arm farm with between 6-14 robots to collect 800K examples for grasping in parallel \cite{levine2016learning}. This  effort demonstrated how exploration can be parallelized and used to collect data at scales like never before. However, apart from scaling up data collection, is there another way to improve performance? In fact, it is known from the rich history of ML that ``all data is not equal''~\cite{sung1994learning, rowley1998neural}. It has been often shown that mining hard examples leads to faster convergence and better performance. Similarly, obtaining better labels for training tends to make learning more efficient and robust. So, when using multiple robots, is there a better way to use multiple robots for exploration leading to faster learning and robust task performance?

\begin{figure}[t!]
\begin{center}
\includegraphics[width=3.3in]{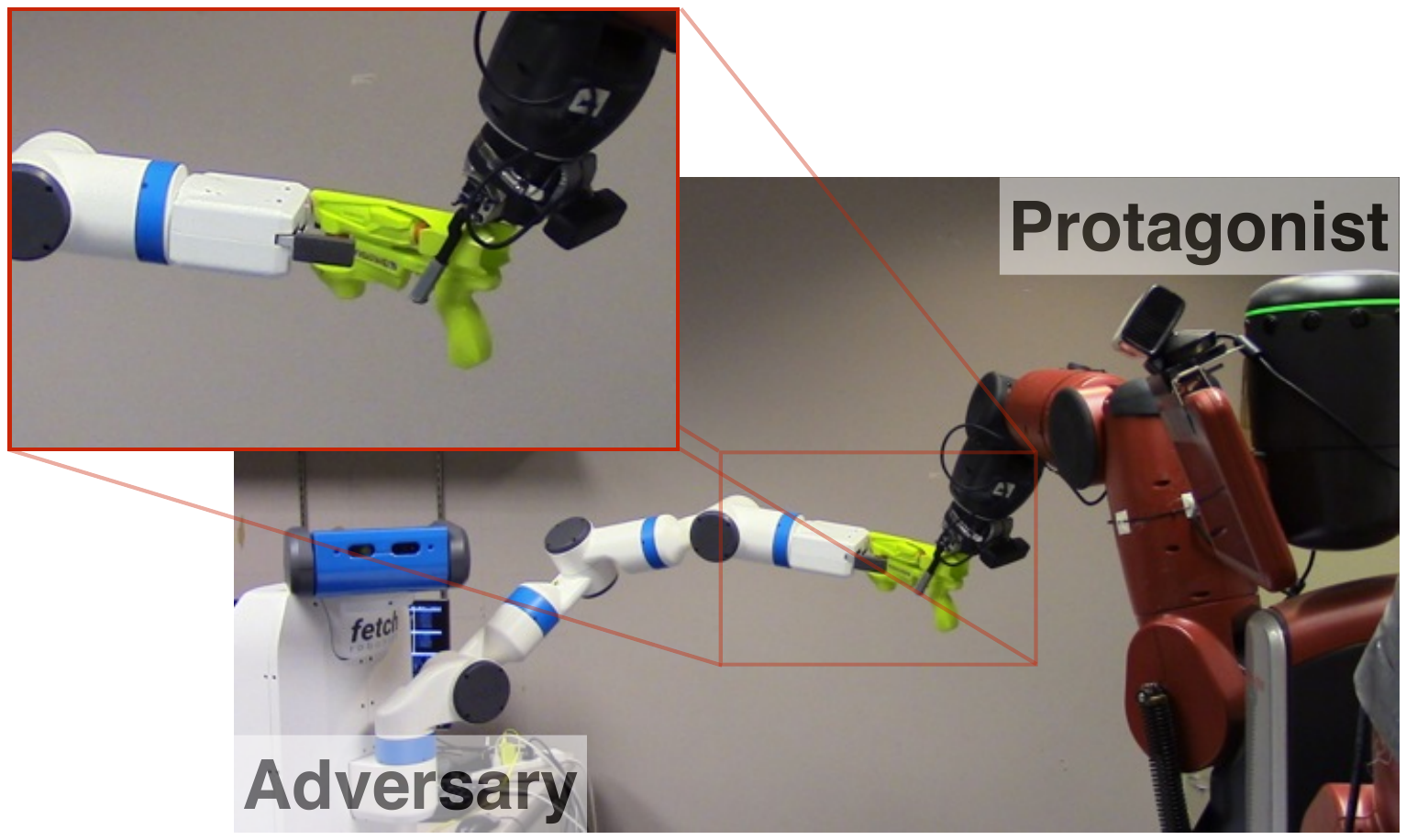}
\end{center}
\caption{We propose an adversarial framework for effective self-supervised learning. In our framework, the protagonist attempts to learn policy for a task such as grasping. While an adversary learns the task to make the protagonist fail at its task. For example, in the figure above, adversary tries to snatch the object from protagonist. Both the policies are learned simultaneously leading to robust learning of protagonist.}
\label{fig:intro_fig}
\end{figure}

Inspired by the recent work in adversarial learning~\cite{goodfellow2014generative, dumoulin2016adversarially}, we propose a novel physical ``adversarial" framework to collect data for manipulation tasks such as grasping. Instead of using robots in a collaborative fashion, we propose to create robot adversaries for self-supervised learning. Specifically, we have an original learner that learns how to perform the task such as grasping; and the adversarial learner tries to learn a task/policy which makes the original learner fail on its task. For example, if the first learner is trying to learn how to grasp the object; the adversarial learner attempts to learn how to steal the object via snatching. This in turn forces the original learner to learn to grasp in a robust manner such that it cannot be snatched by the adversary. In this paper, we create two such adversarial frameworks for grasping. We show how adversarial tasks can help us provide supervision/labeling which rejects weak notions of success leading to faster and better learning.

More importantly, we demonstrate quantitatively that using an adversarial framework leads to a significantly better grasping policy as compared to having two robots collect data in parallel for the same task. 

\begin{figure*}[t!]
\begin{center}
\includegraphics[width=7in]{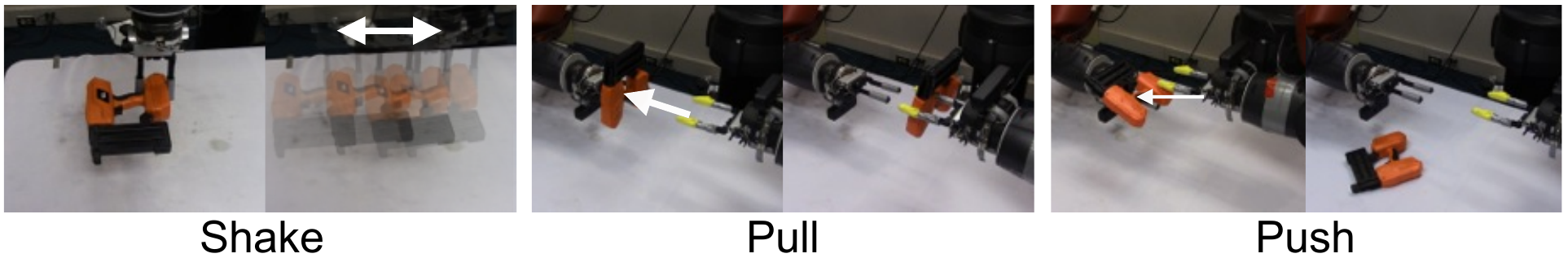}
\end{center}
\vspace{-0.1in}
\caption{Given a weak grasp, an adversary can destabilize in multiple ways. \textbf{Left} shows the motion of a linear shake on the same arm that could destabilize this grasp. Another way to destabilize this is a push grasp on this object by a different arm and then pull. \textbf{Center} shows hows how snatching/pulling can destabilize the grasp, while \textbf{right} shows how the pushing motion can destabilize the grasp.}
\vspace{-0.2in}
\label{fig:framework_fig}
\end{figure*}

%% file: relatedWork.tex
\section{RELATED WORK}

The learning framework we propose intersects with numerous areas of research. In particular, our method employs adversarial learning in physical robots. We then demonstrate the significant improvements this brings to training a robot to grasp.

\noindent{\bf Self-supervised deep learning}.
Data-driven approaches in robotics have gained popularity based on the success of deep neural network methods in other domains including computer vision \cite{krizhevsky2012imagenet} and reinforcement learning \cite{mnih2015human}.  In robotics, to solve data requirements, deep learning methods  have been coupled with self-supervision to generate end-to-end learning solutions  \cite{lenz2015deep, pinto2016supersizing, pinto2016curious, levine2016end, levine2016learning}.  

\noindent{\bf Active Learning}.
Active learning is often employed in domains where data is cheap but labeling is expensive. This is the case in self-supervised paradigms, where the cost of labeling corresponds in part to the time it takes to run the experiment, so there is often a severe limit on the practical amount of data that can be generated. Active learning is motivated by the idea that fewer examples can result in better accuracy if the right data is chosen (refer to \cite{settles09}). In simplified, linear classification problems, it has been shown that exponential decrease is achievable in expectation with no efficiency gains in the worst case \cite{dasgupta2004analysis}. We will show that our technique, which operates over much more complex decision surfaces still achieves significant performance boost over naive sampling.

Our work is also related to Hard-negative/example mining \cite{sung1994learning, rowley1998neural}, formerly referred to as bootstrap learning. Recent methods in computer vision \cite{loshchilov2015online, simo2014fracking, wang2015unsupervised, shrivastava2016training} and sample efficiency in reinforcement learning \cite{schaul2015prioritized} apply hard example mining to train deep neural networks. These methods, however, base their filter of examples by the loss of each sample. In our case, 
 adversarial/antagonist network guides the training of the original network to select harder training set.

The value of more efficient training in robotics has long been investigated due to the high cost of data. Use of active learning has ranged from curriculum learning \cite{sanger1994neural}, selectively filtered uncommon or interesting data \cite{dima2004enabling}, to directed exploration in reinforcement learning (see \cite{thrun1992efficient}).  Much of the recent work in improving exploration for deep learned models has focused reducing uncertainty \cite{houthooft20015variational, osband2016deep} or is novelty seeking \cite{bellemare2016unifying, abel2016exploratory}. Unlike reinforcement learning, our approach is a repeated game of a single time-step.  Moreover, to our knowledge, no previous work in robotics or reinforcement learning has trained an adversary to guide exploration or for sample efficiency.

\noindent{\bf Game Theory}. 
We formulate our adversarial training as a two player zero sum repeated game (refer to \cite{basar1999dynamic} for an overview of game theory). Game theory has been explored in many robotics applications. Multi-agent systems \cite{stone2000multiagent, panait2005cooperative} are often formulated as a cooperative game. In pursuit evasion \cite{lavalle2000robot} and robust control methods \cite{bacsar2008h}, nature acts as the opponent against the system forcing the system to chose safe actions. Our approach is analogous, as the adversary forces the system to choose more stable actions. Unlike existing robotic game theoretic approaches, our technique trains two differing neural networks to act as both players.

\noindent{\bf Adversarial methods}. Generative adversarial methods \cite{goodfellow2014generative, dumoulin2016adversarially} are similar in that they train two neural network agents in a game theoretic settings. However, the objective is to train a model capable of generating examples indistinguishable from training examples. More recent work leverages adversarial examples, similar to hard-negative mining approaches discussed above, to train a more robust classifiers \cite{goodfellow2014explaining}. This method generates new samples based on analysis of a single trained network. Our formulation relies on two discriminative models that adapts to the current behavior of the opposing player.

\noindent{\bf Grasping}.
We test out our technique on the grasping problem, an intensely researched problem which is a complex task requiring a robust solution. Refer to \cite{bicchi2000robotic, bohg2014data} for surveys of prior work. Significant progress has been made recently using data-driven solutions for training deep learned models on both corporeal robots \cite{lenz2015deep, pinto2016supersizing, levine2016end, levine2016learning} and in simulation \cite{kappler2015leveraging, mahler2016dexnet}.  

Early work sought to define stable grasp metrics using analytic methods \cite{ferrari1992planning}. However, later work found such metrics unstable under shaking \cite{balasubramanian2012physical} and pose uncertainty \cite{weisz2012pose, kim2013physically}.  Our approach demonstrates that adversarial agent in a data-driven context leverage these destabilizing factors to train a more effective solution. While we demonstrate the core algorithm on stable grasping, we believe the approach is applicable to other domains.

%% file: approach.tex
\section{Overview}
The goal of this paper is to explore how we can improve self-supervised systems with respect to (a) the quality of supervision and (b) robustness of the trained models. Current self-supervised systems use combinations of sensors to define success/failure on tasks \cite{lenz2015deep, pinto2016supersizing, levine2016end, levine2016learning}. For example, in grasping, a force sensor in the gripper can be used to see if the robot was successful in grasping. However, such supervision is weak and noisy at best. 

In this paper, we argue that learning to defeat an adversarial robot might provide significantly better supervision. Specifically, we learn models for two tasks simultaneously: one model is learned for the original-task such as grasping; the other model learns an adversarial task to defeat the original-task model. As an example, for grasping one adversarial task is to snatch the grasped object from the original model. An important feature of this joint learning is that while the adversarial is learning how to defeat original model; the original model adapts itself to defeat the adversarial model. This leads to greater robustness in the task performance compared to the original model itself. 

Specifically, in this paper, we explore the use of adversarial robots for the task of grasping through two adversarial mechanisms. The first shakes the object in the gripper to break the grasp. The second adversarial mechanism is snatching. Given an object grasped by one arm of a robot, the adversary attempts to snatch the object using a second arm (in our experiment we use Baxter robot, which has two arms). Therefore, the original GraspNet tries to learn the grasping task such that it is robust to either a shaking or snatching adversary, respectively.
Experimentally, we demonstrate that learning via competition from an adversarial model helps improve the performance on the test data. 


\section{Adversarial Learning Framework}
\subsection{Formulation}
For purposes of explanation, the agent is the protagonist being trained to perform the task and the adversary is the antagonist attempting to defeat the agent. Our goal is to learn a non-linear function $\mathcal{P}_{W^p}$ (we use ConvNets and $W^p$ represent the parameters) which given the current state of the environment (represented as $s$) predicts the action parameters $u^p$. Therefore, $u^p=\mathcal{P}_{W^p}(s)$. At the same time, we also try to learn an adversarial task $\mathcal{A}_{W^a}$, which given some state representation($s_+$) after the action $u^p$, predicts the adversarial action parameters $u^a$. Therefore, $u^a=\mathcal{A}_{W^a}(s_+)$. Note that the state of the world $s_+$ after protagonist action, $u^p$, depends on the action and the world.

\begin{figure*}[t!]
\begin{center}
\includegraphics[width=7in]{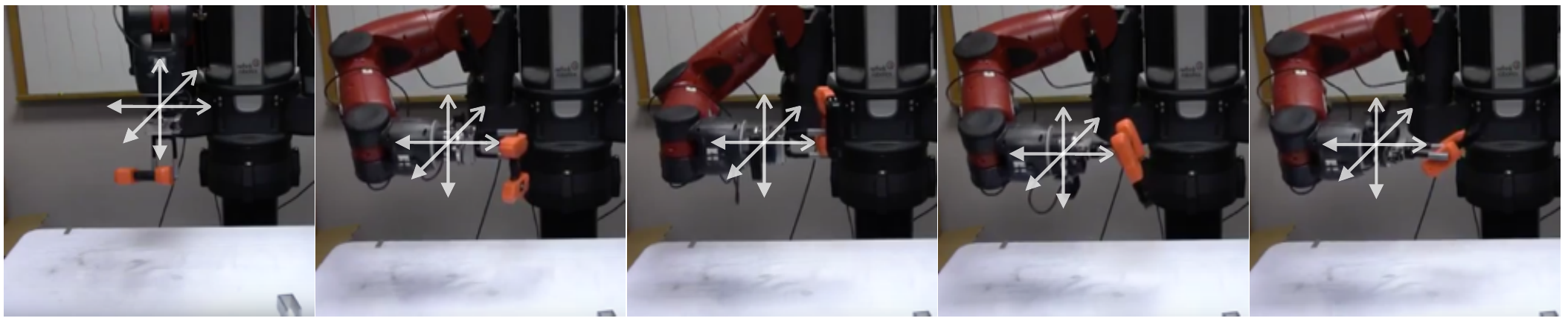}
\end{center}
\caption{The shake space contains 15 discrete actions, with 3 directions of linear shake for 5 different configurations. These 5 configurations are seen as individual images with the 3 directions of shake shown by white arrows.}
\label{fig:shake_space}
\end{figure*}

The optimization goal is to learn $\{W^p,W^a\}$ according to the following criteria:
\begin{equation}
    \min_{W^p}{\left(L^p\left(u^p\right) - \alpha \min_{W^a}{\left(L^a\left(u^a\right)\right)}\right)}
\label{eq:framework}
\end{equation}
where $L^p(u^p)$ is the original task loss and $L^a(u^a)$ is the adversary loss. Therefore, $L^p(u^p)=0$ if the protagonist was successful on the original task and $L^p(u^p)=1$ otherwise. Similarly, $L^a(u^a)=0$ if the adversary was successful in defeating the protagonist. Note that $\alpha$ is the weighting factor for learning of the  protagonist.


The second term in the objective function pushes the adversary to learn parameters $W^a$ such that adversary's action leads to agent's failure (minimizing adversary's loss). However, the original protagonist is trying to learn the parameters $W^p$ such: (a) the agent is able to minimize its own loss (the first term) and (b) maximize adversary loss, which implies the original task is performed so robustly that the adversary cannot defeat it.

\subsection{Optimization}
\noindent {\bf Initializing $\boldsymbol{W^p}$:} We first learn original task, without any adversary in play, only using the weak notion of success. Therefore, we collect the data with random actions and see which actions lead to success in the task using the sensors. This data is used to learn an original ConvNet. Note that the ConvNet $\Phi_{W^p}$ gives probability of success for all possible actions and therefore the policy is to select the action in greedy manner.

\begin{equation}
    \mathcal{P}_{W^p}(s) = \arg\max_{u}{\Phi_{W^p}(s,u)}
\end{equation}

\noindent {\bf Initializing $\boldsymbol{W^a}$:} Given the initial learned ConvNet for the original task, we use it to perform the task followed by an adversary action $u_a$. Initially, we use a random adversary policy and collect data to observe effect of random adversary actions on the original task. We use the collected data to learn ConvNet for initial adversary task. Again, the ConvNet $\Phi_{W^a}$ gives probability of success for all possible adversary actions.

\noindent {\bf Joint Training:} Once we have initialized both protagonist and adversary task networks, we use the ConvNets to collect joint data. The original task network ($\Phi_{W^p}$) defines the probability of success for all actions, we select the action with the highest probability of success and perform the action $u_p$. If the $u_p$ is unsuccessful, we update parameters of original task network $W^p$. If the $u_p$ is successful, then the adversary network ($\Phi_{W^a}$) is used to select an adversary action $u_a$. Based on the success or failure of adversary action, both the adversary and protagonist network parameters are updated. This procedure is repeated by generating a series of batch updates for each iteration.  Training repeated multiple epochs until a minimum accuracy threshold on the train set was achieved.

\begin{figure*}[t!]
\begin{center}
\includegraphics[width=5in]{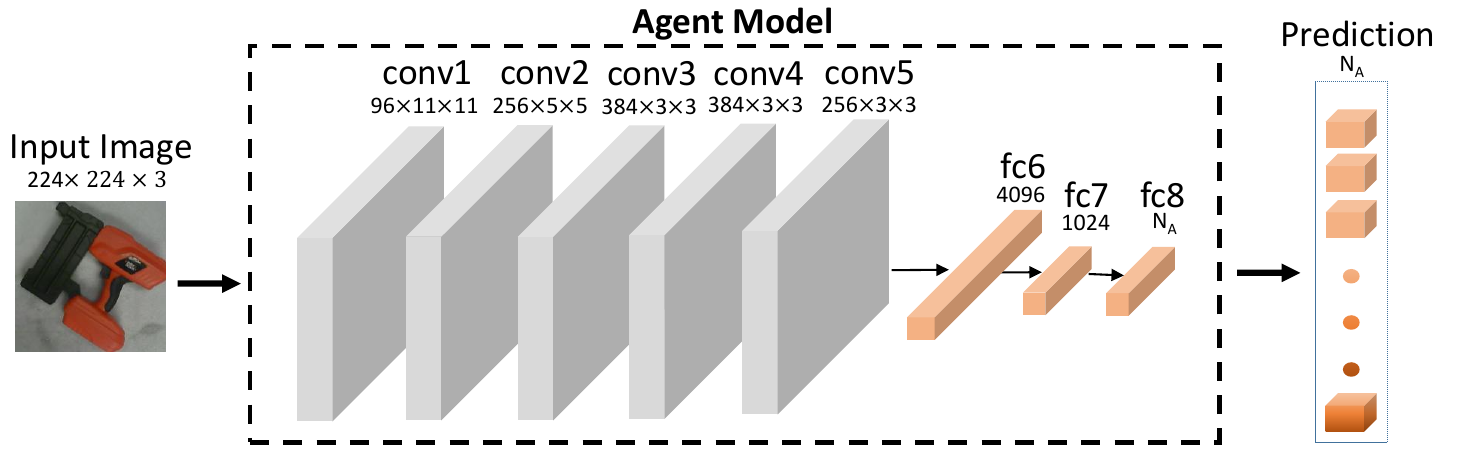}
\end{center}
\caption{The network used by both the protagonist and the antagonist is modeled after AlexNet \cite{krizhevsky2012imagenet}. The output of this network is scaled by using a sigmoidal function.}
\label{fig:agent}
\end{figure*}

\section{Experimental Framework}
We demonstrate the effectiveness of the proposed adversarial framework for the task of grasping using two different adversaries: shaking the hand holding the object; snatching the object using the other hand.

\subsection{Grasp prediction formulation} 
\begin{figure*}[t!]
\begin{center}
\includegraphics[width=6.0in]{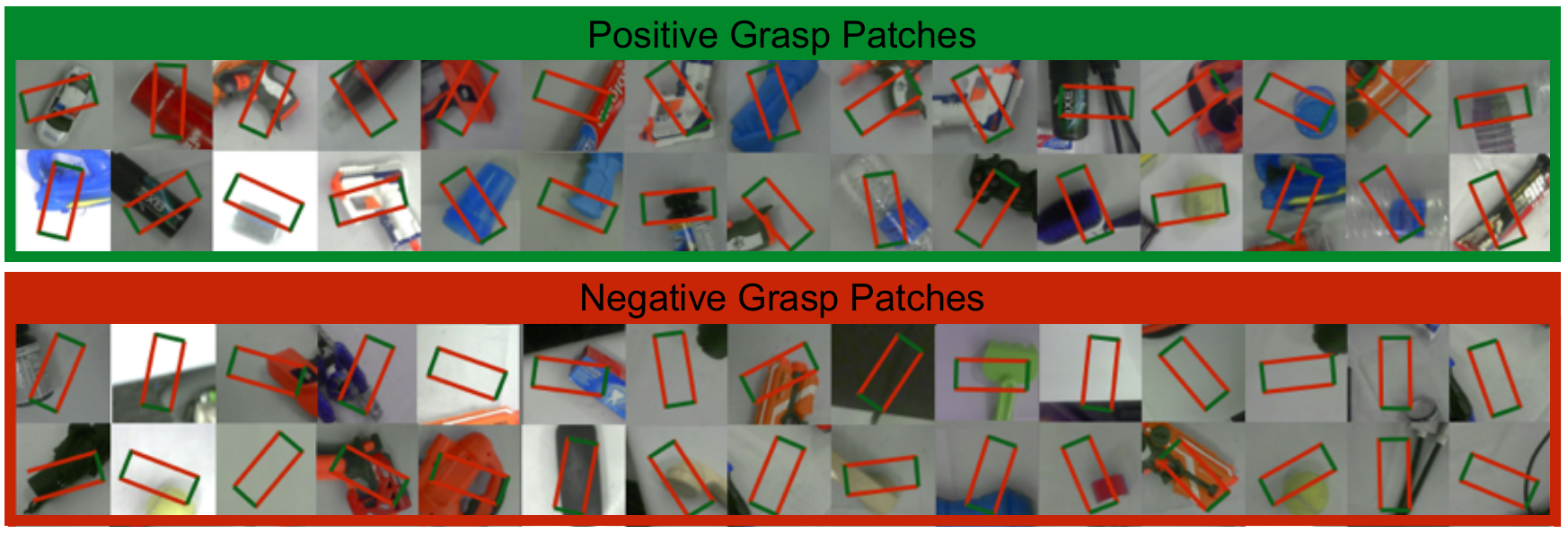}
\end{center}
\caption{Examples of successful (top) and unsuccessful grasps (bottom). We use a patch based representation for grasping: given an input patch we predict 18-dim vector which represents whether the center location of the patch is graspable at $0^{\circ}$, $10^{\circ}$, \dots $170^{\circ}$.}
\label{fig:grasp_data}
\end{figure*}

We formulate our problem to perform planar grasping~\cite{pinto2016supersizing}, which defines action space/grasp configuration using three parameters: $(x,y,\theta)$ as position of grasp point on the surface of table and angle of grasp, respectively. This reduces the task of finding successful grasps to finding a successful configuration, $(x_s,y_s,\theta_s)$ in a given image $I$. Examples of planar grasps are shown in Figure~\ref{fig:grasp_data}. However, as mentioned in \cite{pinto2016supersizing}, this formulation is problematic due to the presence of multiple grasp locations for each object. Therefore, we sample image patches from the input image and evaluate them by the ConvNet to give probability of success for all the grasp angles with grasp location being center of patch.
Given an image patch, we output an 18-dimensional likelihood vector where each dimension represents the likelihood of whether the center of the patch is graspable at $0^{\circ}$, $10^{\circ}$, \dots $170^{\circ}$. Therefore, the grasping problem can be thought of as 18 binary classification problems. 

\subsection{Grasping as the protagonist agent}
Figure~\ref{fig:agent} with $N_a=18$ describes the network architecture for our protagonist agent policy. Given an input image $I$, multiple ($N_g$) patches are sampled. For each of these patches $I_g$ the agent network predicts the probabilities of successful grasping in the 18 different grasp angles. This gives a $N_g\times N_a$ grasp probability matrix. Depending on the exploration strategy, an element $(I_g,N_a)$ is chosen from this matrix which corresponds to the grasp configuration $(x_g,y_g,\theta_g)$. This grasp is then executed. The agent policy $\mathcal{P}_{w^P}(s)$ hence uses the state representation as the input image, i.e. $s \equiv I$ and outputs the grasp action $u^p \equiv (x_g,y_g,\theta_g))$. 

\vspace{2pt}

\subsection{Shake formulation}
\begin{figure}[t!]
\begin{center}
\includegraphics[width=3.5in]{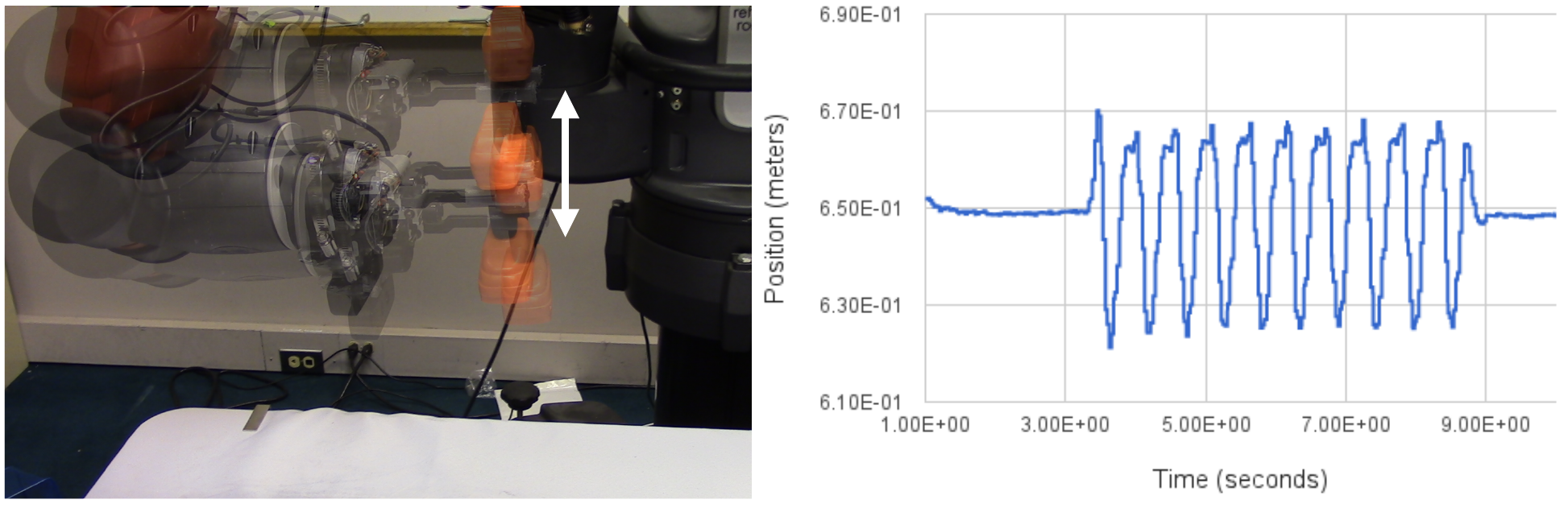}
\end{center}
\caption{\textbf{Left} shows an example of the shake motion. \textbf{Right} shows the end effector position trajectory that shakes the grasped object at 2Hz with an amplitude of 25mm}
\label{fig:shake_formulation}
\end{figure}

Shaking is used as one of the adversarial mechanisms (Figure ~\ref{fig:shake_formulation}). After the protagonist grasping agent grasps an object, the antagonist attempts to destabilize the grasp by vigorously shaking the end effector. The shake action space is discrete with 15 possible options. Each shake action corresponds to pair of end effector orientation and direction of linear shake. There are 5 end effector orientations used in this work as shown in Figure~\ref{fig:shake_space} and 3 possible directions of linear shake for each of these orientations. A sample end effector shake motion can be seen in Figure ~\ref{fig:shake_formulation}. Note that the frequency and amplitude of the shake is held constant for all the shake actions.

\subsection{Snatch formulation}
\begin{figure}[t!]
\begin{center}
\includegraphics[width=3.4in]{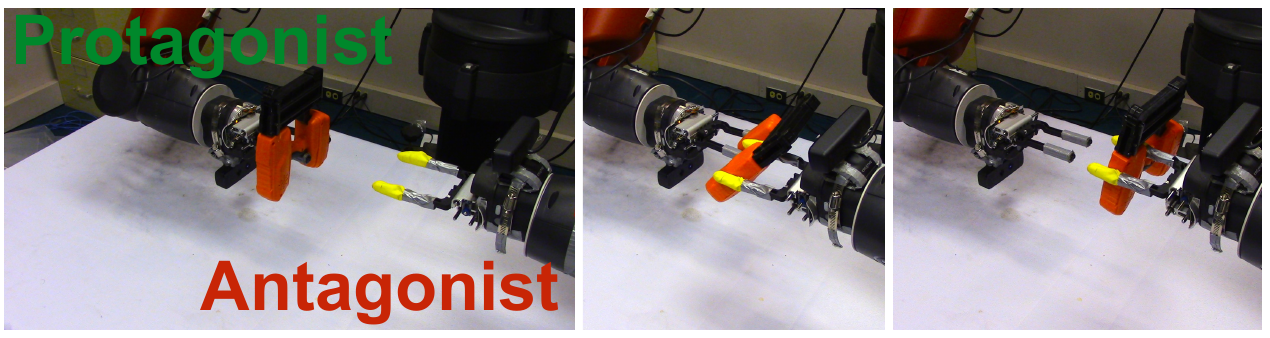}
\end{center}
\caption{During a snatch action, the adversary (\textbf{left}) first performs a push grasp on the object held by the other arm (\textbf{center}) and then then pulls (\textbf{right}).}
\label{fig:pull_formulation}
\end{figure}

\begin{figure}[t!]
\begin{center}
\includegraphics[width=2in]{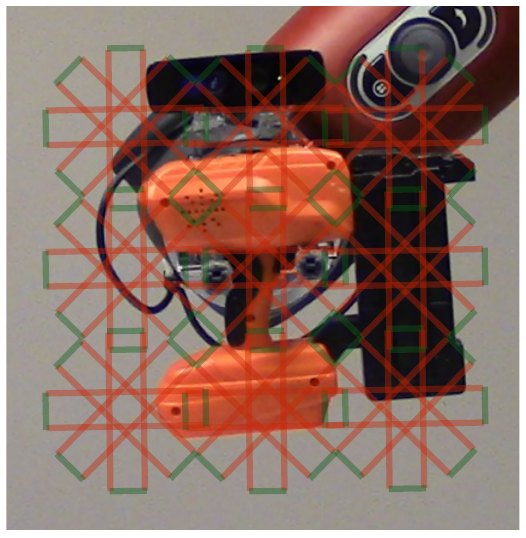}
\end{center}
\caption{The snatch space contains 36 discrete actions, with 9 grasp points and 4 angles of grasp on each of these grasp points. All of these 36 possible grasps are overlayed on an image of the grasped image in this figure}
\label{fig:pull_space}
\end{figure}

The other adversarial mechanism we explore is snatching (Figure ~\ref{fig:pull_formulation}). After the protagonist grasping agent grasps an object, the antagonist attempts to destabilize the grasp by attempting first to push grasp ~\cite{dogar2011framework} followed by pulling the object. Note that here the antagonist does not have control over the protagonist's arm but has control over a second arm that attempts this push grasp. The snatch action space is discrete with 36 possible options. Each snatch action corresponds to a grasp point in the plane of the object (see Figure ~\ref{fig:pull_space}). There are 9 translational configurations and 4 rotational configurations. A sample end effector pull motion can be seen in Figure ~\ref{fig:pull_formulation}.

\subsection{Antagonist adversary}
 Given the original state $s$ and the executed action $u^p$, we represent the current state $s_+$ as a function of $s$ and $u^p$ (say $\mathcal{F}(s,u^p)$). In this work $\mathcal{F}(s=I,u^p=(x_G,y_G,\theta_G))$ is the image patch of $I$ centered at $(x_G,y_G)$ an rotated by $-\theta_G$. Hence, $s_+$ is the grasped patch of the object rotated by the angle of grasp so that it is visible in canonical viewpoint. This effectively encodes the image patch after the agent's fingers grasp the object. This means that the adversary's state representation $s_+$ is also the image $I^A$.

 The adversary policy $\mathcal{A}_{W^A}(s_+)$ now has to predict the adversarial action $u^a$. For this we again use the deep network ($\Phi_{W^A}$) architecture in Figure ~\ref{fig:agent}. For shaking, there are 15 actions, so $N_A=15$ and $u^a \in {1,2,3,..,15}$. Snatching has 36 actions, so $N_A=36$ and $u^a \in {1,2,3,..36}$. Given the state $s_+\equiv I^A$, the network produces the probabilities of the adversary succeeding for the $N_A$ different actions. Based on the adversary's exploration strategy, the action $u^a$ is sampled from the probability distribution and executed. The adversary policy $\mathcal{A}_{W^A}(s_+)$ hence uses the state representation as a rotated patch and outputs the adversarial action $u^a$. 

\begin{figure*}[t!]
\begin{center}
\includegraphics[width=6.5in]{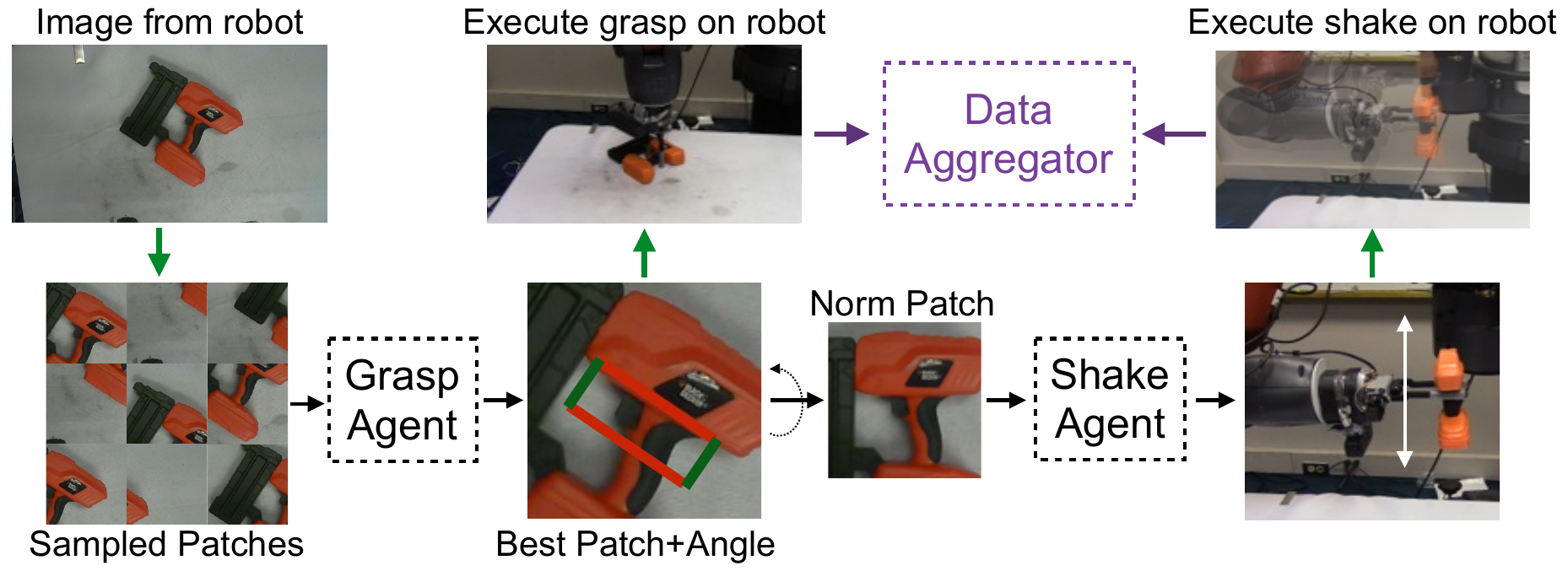}
\end{center}
\caption{The shaking adversarial framework' pipeline is shown here. Given an image, the grasping agent inputs patch samples and outputs the best prediction for grasp. If it is successful in grasping the object, the shaking agent chooses its best prediction to destabilize the grasp and executes it. All of this data is aggregated for training the next iteration of the grasping and shaking agents.}
\label{fig:grasp_with_shake}
\end{figure*}

\subsection{Network architecture} For all the networks, we used the same ConvNet architecture. Specifically, the architecture used in this work (Figure ~\ref{fig:agent}) is modeled on AlexNet ~\cite{krizhevsky2012imagenet}. The only difference being the fewer neurons in fc7 (1024 vs 4096). The output of the network is scaled to (0,1) using a sigmoidal response function. The convolutional weights are initialized from ImageNet pretrained weights, while the fully connected weights are initialized randomly.

\subsection{Learning to grasp with shake adversary}
In our formulation to jointly train the grasping agent and the shaking adversary, we would first initialize the policy of the agent and the adversary. For the grasping task, we
first collect random samples \cite{pinto2016supersizing} and train an initial model $(\mathcal{P}^{0}_w )$ on this. Let this dataset of random grasps be called $D^{0}_G$. Each element of $D^{0}_G$ contains
a patch grasped $I_G$, the discrete angle at which it was grasped $\theta_D$ and if the grasp succeeded or failed. The training target $y^{\prime}_{\theta_D}$ is given by
\begin{equation}
    y^{\prime}_{\theta_D}= 
\begin{cases}
    0,& \text{if grasp failed}\\
    1,& \text{if grasp succeeded}
\end{cases}
\end{equation}

Using this, the grasping network is trained using the sigmoidal outputs of the network in $\mathcal{P}^{0}_{w}$ as predictions $y_{\theta_D}$. The loss for this training is the binary cross entropy loss between $y_{\theta_D}$ and $y^{\prime}_{\theta_D}$. This network is trained using RMSProp~\cite{tieleman2012lecture}.

Now given the initial grasping policy $\mathcal{P}^{0}_{w}$, we grasp objects by sampling from the probability distribution generated by the network in $\mathcal{P}^{0}_{w}$. Once this grasp is executed random shakes are applied on the grasped objects. We call this dataset $\{D^1_G,D^1_S\}$, where $D_S$ is the shake data (success or failure of the shaking adversary). 

Given a successful grasp $(I_G,\theta_D)$ in $D^1_G$, first $I_G$ is rotated with the angle represented by $\theta_D$ to get $I^A$. The target for the network in $\mathcal{A}^{1}_{W^A}$ is now

\begin{equation}
    y^{\prime}_{shake}= 
\begin{cases}
    0,& \text{if shake failed in destabilizing grasp}\\
    1,& \text{if shake succeeds and causes object to fall}
\end{cases}
\end{equation}

Using this, the shaking network is trained using the sigmoidal outputs $\Phi^{1}_{W^A}$ of the network in $\mathcal{A}^{1}_{w}$ as predictions $y_{shake}$. The loss for this training is the binary cross entropy loss between $y_{shake}$ and $y^{\prime}_{shake}$. This network is trained using RMSProp. Hence using these random shakes, the initial shaking policy $\mathcal{A}^{1}_{W^A}$ is learnt.

To train $\mathcal{P}^{1}_{w}$, we again use the dataset $\{D^1_G,D^1_S\}$. The training target $y^{\prime}_{\theta_D}$ for $\mathcal{P}^{1}_{w}$, is given by 
\begin{equation}
    y^{\prime}_{\theta_D}= 
\begin{cases}
    0,& \text{if grasp failed}\\
    1 - \alpha\cdot\max_{u}{\Phi^{1}_{W^A}(s_+)},              & \text{if grasp succeeded}
    
\end{cases}
\end{equation}

Here, $\max_{u}{\Phi^{1}_{W^A}(s_+)}$ is the maximum probability the network in $\mathcal{A}^{1}_{W^A}$ believes it can destabilize the grasping agent. $\alpha$ is a factor that controls how strong the adversary controls the learning of the agent. Using these labels, network in $\mathcal{P}^{1}_{w}$ is trained.

Once we have our initial models $\mathcal{P}^{1}_{w}$ and $\mathcal{A}^{1}_{W^A}$, we now follow the iterative training process. For iteration $i$, we grasp objects using $\mathcal{P}^{i}_{w}$ with importance sampling on grasp probabilities. For successful grasps shakes are applied according to $\mathcal{A}^{i}_{W^A}$ with greedy sampling on shake probabilities. The data collected $\{D^{i+1}_G,D^{i+1}_S\}$ is then used to train $\mathcal{A}^{i+1}_{W^A}$ first followed by $\mathcal{P}^{i+1}_{W^P}$.
The data collection procedure is described in Figure ~\ref{fig:grasp_with_shake}.

%% file: results.tex
\section{DISCUSSION}

\subsection{Hardware Setup}
All our experiments were run on a Baxter robot. The protagonist arm is employed with parallel jaw grippers which can apply up to 35N of gripping force with maximum payload of 2.2Kg. However, for the collection of adversarial data, the gripping force was reduced to 7N to bias the framework towards the adversary. However for final testing we report results with the maximum gripping force as well as with the train time gripping force. For experiments with the snatching adversary, the other arm of Baxter is used to snatch grasped objects from the protagonist arm.

\subsection{Data Collected}
The grasping model was initialized using 40K grasping tries from   ~\cite{pinto2016supersizing}. 

For the shaking task, a total of 9k grasps are attempted over 3 iterations of batch adversarial learning. Out of these, the protagonist successfully grasped 2.4k. This low grasping rate can be attributed to our exploration strategy where we use importance sampling to sample grasps. Out of these successful grasps 0.5k were dislodged by the adversary.

For the snatching task, a total of 2k grasps are attempted over 2 iterations of batch adversarial training. The grasp network in this case was initialized with the grasp network trained by the shake adversary. Out of the 2k grasps attempted, the protagonist successfully grasped 0.7k. Of these, 0.2k adversarial snatches were successful in dislodging the protagonist.

For the baseline, we trained a grasping model using 56K training examples. Note that the baseline is trained with more data points assuming two robots can collect data faster in parallel (collaborative setting rather than adversarial).

\subsection{Grasping Results}

\begin{figure}[t!]
\begin{center}
\includegraphics[width=3.3in]{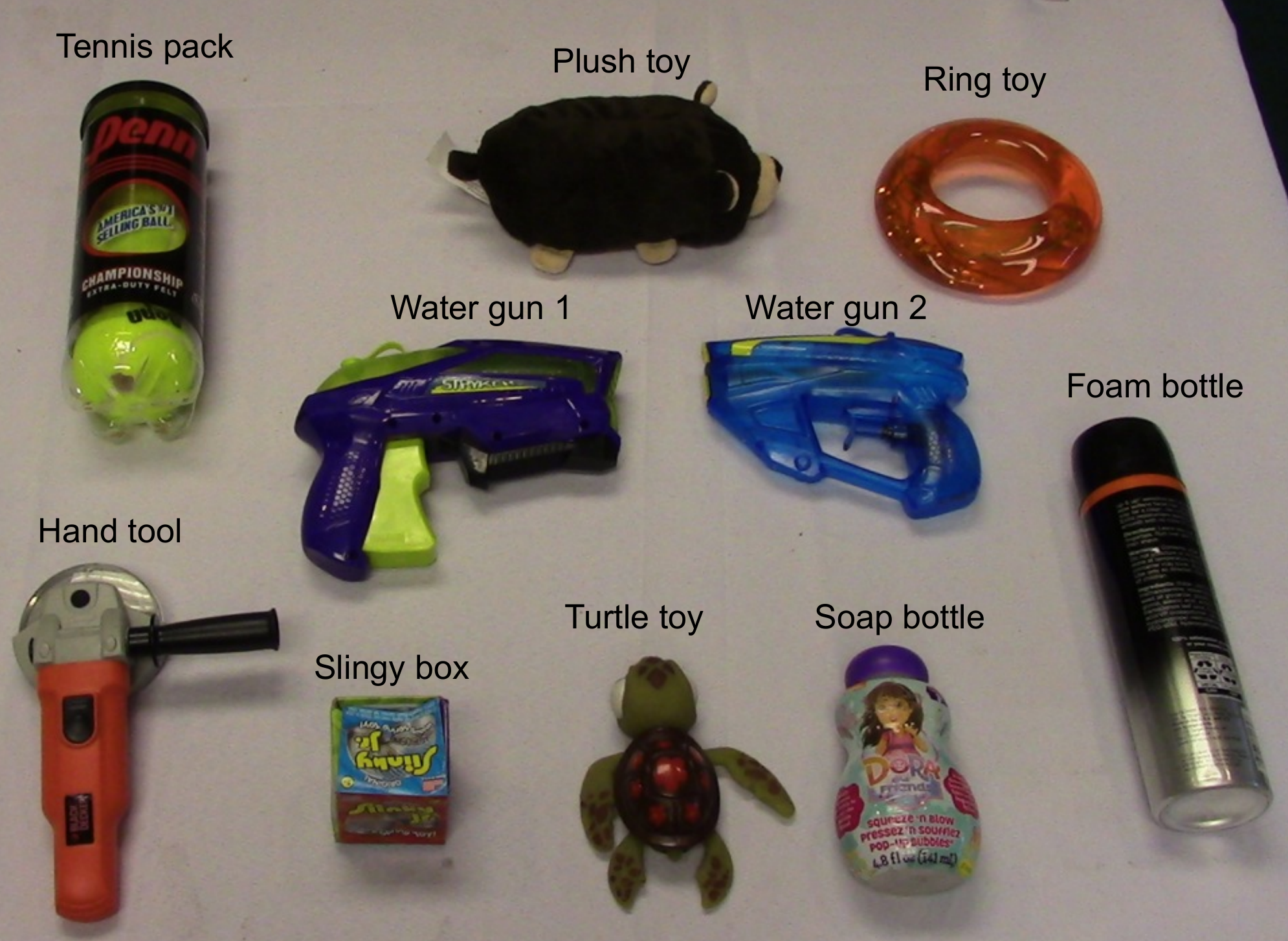}
\end{center}
\caption{We perform our grasping analysis on these 10 objects. Note that the objects present a mix of difficulty, with deformable objects, non convex objects and heavy objects.}
\label{fig:objects}
\end{figure}

For demonstrating the improvements of our adversarial framework, we use the objects shown in Figure ~\ref{fig:objects}. These objects are novel and have not been seen by the robot before. As a baseline we train a grasping model with the same amount of data as the adversarial model but without adversarial supervision. This is similar to ~\cite{pinto2016supersizing}.
The results for grasping with low gripping force (20\% of maximum gripping force), without additional grip and by sampling only 128 patches on the input image can be seen in Table ~\ref{tab:results_low}. For a successful grasp to be reported, the object must be grasped and lifted 30cm from the table. 

We can clearly see the improvement by the adversarial framework among these objects. After 3 iterations of training with shaking adversary, our grasp rate improves from 43\% to 58\%. Note that our baseline network that does not perform adversarial training has a grasp rate of only 47\%. This clearly indicates that having extra supervision from adversarial agent is significantly more useful than just collecting more grasping data. What is interesting is the fact that 6K adversary examples lead to 52\% grasp rate (iteration 1) where as 16K extra grasp examples only have 47\% grasp rate. This clearly shows that in case of multiple robots, training via adversarial setting is a more beneficial strategy.

Finally, we used the model trained after competing with shaking adversary to further train it by competing against a snatching adversary. Adding iterations of adversarial training with the snatching adversary further improves the performance to 65\%.

To stay closer to the baseline ~\cite{pinto2016supersizing}, we increase the grasping force to the maximum possible and attach a rubber grip to the fingers similar to the one in ~\cite{pinto2016supersizing}. Instead of sampling 128 patches, 10 times more patches are sampled to ensure better coverage of the image space. The results for this can be seen in Table ~\ref{tab:results_high}. Once again, the shaking adversarial framework (80\%) beats the baseline (68\%) handily. In the case of adding snatching adversary (82\%), we see a small 2\% improvement.

\begin{table*}[]
\centering
\caption{Grasping success (out of 10 tries) over iterations with low gripping force and few samples}
\label{tab:results_low}
\begin{tabular}{l|c|ccc|cc|}
\multicolumn{1}{c|}{\multirow{3}{*}{Object}} & \multicolumn{1}{c|}{\multirow{3}{*}{Grasping \textbf{without} adversary}} & \multicolumn{5}{c|}{Grasping \textbf{with} adversary} \\ \cline{3-7} 
\multicolumn{1}{c|}{} & \multicolumn{1}{c|}{} & \multicolumn{3}{c|}{Shake Adversary} & \multicolumn{2}{c|}{Snatch Adversary} \\ \cline{3-7} 
\multicolumn{1}{c|}{} & \multicolumn{1}{c|}{} & \multicolumn{1}{c}{Iteration-0} & \multicolumn{1}{c}{Iteration-1} & \multicolumn{1}{c|}{Iteration-2} & \multicolumn{1}{c}{Iteration-0} & \multicolumn{1}{c|}{Iteration-1} \\ \hline
Tennis Pack & 10 & 10 & 10 & 10 & 10 & 10 \\
Hand Tool & 4 & 3 & 3 & 3 & 4 & 3 \\
Water Gun 1 & 4 & 3 & 5 & 8 & 7 & 8 \\
Slingy Box & 5 & 6 & 6 & 6 & 6 & 7 \\
Plush Toy & 4 & 0 & 3 & 4 & 1 & 5 \\
Turtle Toy & 2 & 2 & 3 & 2 & 1 & 3 \\
Water Gun 2 & 4 & 4 & 4 & 7 & 8 & 9 \\
Soap Bottle & 6 & 7 & 9 & 9 & 8 & 10 \\
Ring Toy & 0 & 0 & 0 & 0 & 0 & 0 \\
Foam Bottle & 8 & 8 & 9 & 9 & 10 & 10 \\ \hline
\multicolumn{1}{c|}{Overall} & \textbf{47\%} & 43\% & 52\% & \textbf{58\%} & 55\% & \textbf{65\%}
\end{tabular}
\end{table*}

\begin{table}[]
\centering
\caption{Grasping success (out of 10 tries) with high gripping force and more samples}
\label{tab:results_high}
\begin{tabular}{l|c|c|c|}
\multicolumn{1}{c|}{\multirow{2}{*}{Object}} & \multicolumn{1}{c|}{\multirow{2}{*}{Grasping \textbf{without} adversary}} & \multicolumn{2}{c|}{Grasping \textbf{with} adversary} \\ \cline{3-4} 
\multicolumn{1}{c|}{} & \multicolumn{1}{c|}{} & \multicolumn{1}{c|}{Shake} & \multicolumn{1}{c|}{Snatch} \\ \hline  
Tennis Pack  & 6 & 10 & 10 \\
Hand Tool  &10 & 6 & 6 \\
Water Gun 1  & 5 & 8 & 8 \\
Slingy Box &  8 & 8 & 8 \\
Plush Toy &  2 & 6 & 6 \\
Turtle Toy &  4 & 4 & 6 \\
Water Gun 2 & 3 & 8 & 8 \\
Soap Bottle &  10 & 10 & 10 \\
Ring Toy & 10 & 10 & 10 \\
Foam Bottle &  10 & 10 & 10 \\ \hline
\multicolumn{1}{c|}{Overall} & \textbf{68\%} & \textbf{80\%} & \textbf{82\%}
\end{tabular}
\end{table}

%% file: conclusion.tex
\section{CONCLUSION}
Data-driven approaches in robotics have demonstrated significant improvements in recent years  \cite{lenz2015deep, pinto2016supersizing, pinto2016curious, levine2016end, levine2016learning}.  However, these recent approaches are data hungry, which often limits their applicability. We presented an adversarial self-supervised learning framework to help overcome this data issue. By pitting an adversary against an agent attempting to learn the core task, an agent is forced to learn robust solutions resulting in better and more efficient learning. We tested our approach on grasping problems to validate and evaluate its benefits.  The result is a significant improvement over baseline in grasping of novel objects: an increase in overall grasp success rate to 82\% (compared to 68\% if no adversarial training is used).  Even more dramatically, if we handicapped the grasping by reducing maximum force and contact friction, the method achieved 65\% success rate (as compared to 47\% if no adversarial training was used).